\documentclass[conference,compsoc]{IEEEtran}

\usepackage{cite}
\usepackage{amsmath,amssymb,amsfonts}
\usepackage{booktabs}
\usepackage{graphicx}
\usepackage{multirow}
\usepackage{placeins}
\usepackage{url}

\IEEEoverridecommandlockouts

\begin{document}

\bstctlcite{BSTcontrol}

\title{MyoSem: Aligning Electromyography with Natural-Language Action Semantics}

\author{\IEEEauthorblockN{Chiyue Wang, Yang Gao, and Zhanpeng Jin\textsuperscript{*}}
\IEEEauthorblockA{School of Future Technology, South China University of Technology, Guangzhou 511442, China}
\thanks{\textsuperscript{*}Corresponding author: zjin@scut.edu.cn}}

\maketitle

\begin{abstract}

Electromyography (EMG) records muscle activation for hand gesture recognition, prosthetic control, and wearable interaction. Existing EMG methods usually formulate hand action understanding as prediction over fixed gesture labels, leaving EMG signals and action-semantic text unaligned in a common feature space and limiting bidirectional querying between signals and hand-action descriptions. We present MyoSem, an EMG--action semantic alignment framework that maps hand EMG and multi-view natural-language action descriptions into a shared feature space for retrieval. MyoSem builds semantic prototypes from multi-view descriptions, encodes EMG with local activation-aware representations, and uses semantic queries to aggregate local EMG evidence for bidirectional EMG--text retrieval. We evaluate MyoSem under seven protocols on NinaPro DB2, EMG2Pose, and NinaPro DB2$\rightarrow$DB3, covering random splits, user shifts, held-out action classes, and able-bodied-to-amputee transfer. Across this evaluation, MyoSem achieves the strongest overall bidirectional retrieval performance among the evaluated methods. Ablations and representation analyses show how semantic prototypes, activation-aware encoding, and semantic queries shape the learned space. Overall, MyoSem moves EMG hand action understanding from fixed-label recognition toward language-queryable semantic retrieval.

\end{abstract}


\section{Introduction}
\label{sec:introduction}

Electromyography (EMG) records muscle activity for gesture recognition, prosthetic control, and wearable interaction \cite{atzori2014ninapro,kaifosh2025neuromotor}. Existing systems usually predict task-defined gesture labels \cite{duan2021capgmyo,yang2024emgbench}. This closed-label formulation leaves EMG representations and action-semantic descriptions, such as finger involvement, motion progression, and final hand shape, unaligned in a shared feature space. Descriptions therefore cannot directly query EMG segments, and EMG segments cannot retrieve semantically matching descriptions. We study bidirectional retrieval between hand EMG and multi-view action descriptions.

\begin{figure}[t]
  \centering
  \includegraphics[width=0.92\columnwidth]{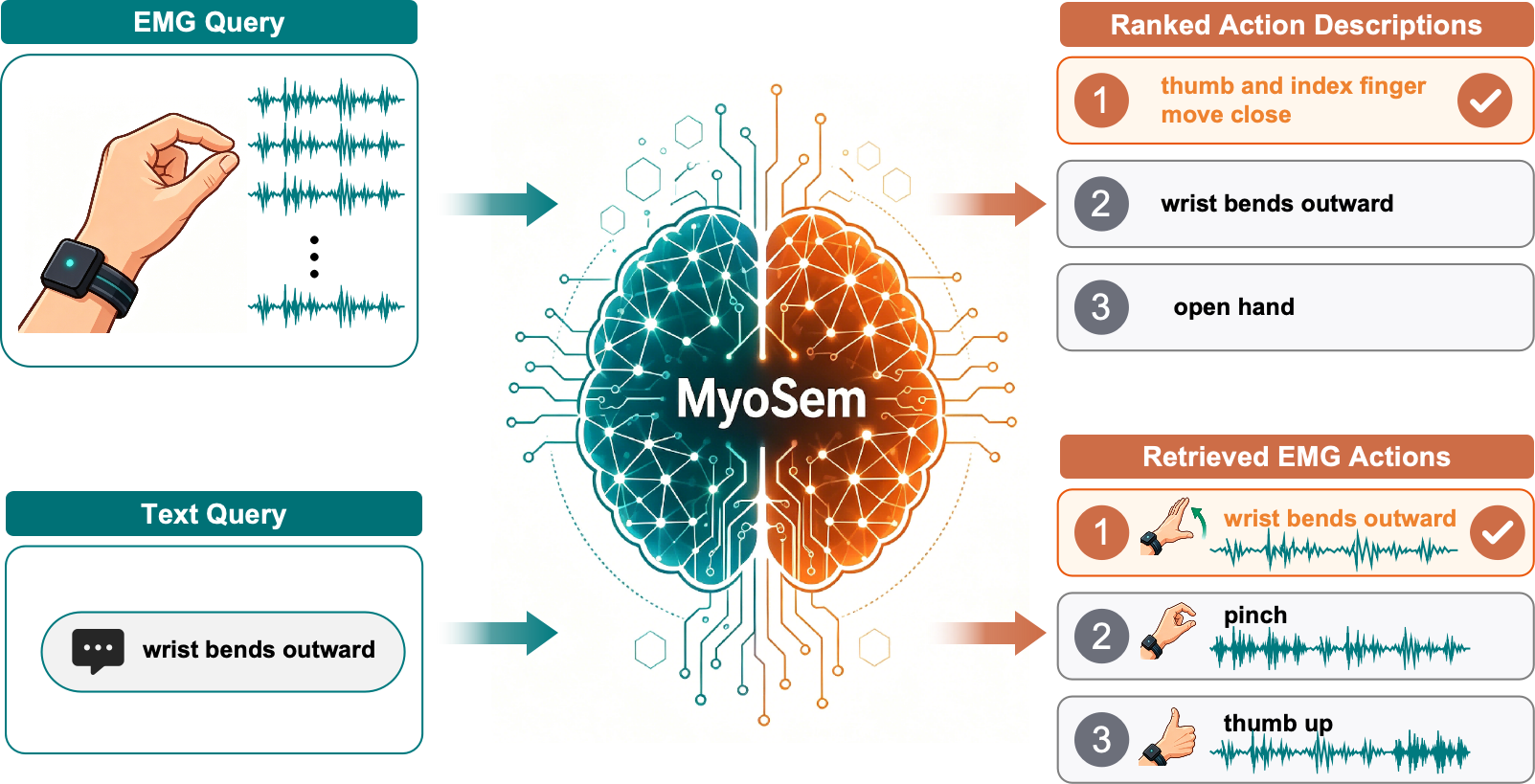}
  \caption{MyoSem as a bidirectional EMG--language retrieval interface. Given an EMG segment, the model ranks multi-view action descriptions; conversely, an action description retrieves matching EMG samples from a gallery.}
  \label{fig:myosem_teaser}
\end{figure}

Representation learning improves EMG features through pretraining, cross-user modeling, and pose- or kinematics-based supervision \cite{huang2026aemg,cui2025cpep,cui2026embridge}. EMG--language systems address unvoiced-speech transcription, medical analysis, and language-model-assisted recognition \cite{emg2text2025,emgllm2025,semgllm2025,wang2026llmsemg}. MyoSem aligns hand EMG with multi-view action descriptions and evaluates EMG--description correspondence through bidirectional retrieval.

This goal creates three design requirements. Text representations must capture finger involvement, progression, and final configuration while separating neighboring actions. EMG representations must preserve local amplitude, energy, and temporal-variation cues that vary across people and acquisition conditions \cite{yang2024emgbench,hudgins1993emgfeatures}. Because continuous EMG and discrete text also differ in granularity, alignment must aggregate local EMG evidence for action prototypes \cite{liang2022mindgap}. MyoSem addresses these requirements with multi-view semantic organization, activation-aware EMG encoding, and semantic-query alignment, allowing EMG segments and action descriptions to act as mutual queries.

We evaluate seven protocols across NinaPro DB2, EMG2Pose, and DB2$\rightarrow$DB3, covering in-distribution retrieval, user shifts, held-out EMG classes with descriptions introduced at inference, and able-bodied-to-amputee transfer. MyoSem achieves the best EMG-to-text R@3 in all seven protocols and stronger text-to-EMG retrieval than all evaluated text encoders; ablations and representation analyses examine the three design choices.

We contribute: (1) a shared-space formulation of EMG hand-action understanding, evaluated by bidirectional retrieval under user, action-class, and able-bodied-to-amputee shifts; (2) MyoSem, which combines multi-view semantic organization, activation-aware encoding, and semantic-query alignment; and (3) a seven-protocol evaluation with ablations and representation analyses.

\section{Related Work}
\label{sec:related-work}

\subsection{Language Alignment for Motion, Sensors, and Biosignals}

Cross-modal systems learn reusable non-text representations and map them into text or unified multimodal spaces. CLIP and ImageBind show that shared semantic spaces support cross-modal matching and retrieval \cite{radford2021clip,girdhar2023imagebind}. This paradigm has extended to IMU, motion sequences, human activity, and EEG \cite{xu2021limubert,zhu2023motionbert,haresamudram2024harfm,jiang2024labram}; related work maps these signals into text or shared spaces \cite{tevet2022motionclip,skeletonclip2024,eegclip2024,moon2023imu2clip,sensorlm2025,miao2024goat} and builds language-model interfaces for motion and activity understanding \cite{jiang2023motiongpt,chen2024sensor2text,li2024sensorllm,yan2025lanhar}. Motion and IMU record visible movement more directly, whereas hand EMG records the muscle activation producing it. MyoSem organizes this physiological evidence with multi-view action descriptions for retrieval.

\subsection{EMG Representation Learning and Language Interfaces}

EMG representation learning has used self-supervision, masked modeling, and large-scale pretraining for cross-setting generalization \cite{reactemg2025,huang2026aemg}. Pose and kinematics provide structured supervision for mapping EMG to joint or trajectory structures \cite{cui2025cpep,cui2026embridge,lin2021bertsEMG,liu2021neuropose,salter2024emg2pose}. Text and language models support unvoiced-speech transcription, medical reports, and action recognition \cite{emg2text2025,emgllm2025,semgllm2025,wang2026llmsemg}. MyoSem uses multi-view descriptions as semantic prototypes and learns bidirectional retrieval between EMG and those prototypes. Activation-aware encoding retains local muscle-activation evidence, and semantic-query alignment aggregates it for prototype alignment.

\section{Method}
\label{sec:method}

MyoSem aligns hand EMG and action descriptions through the following three components.

\begin{figure*}[t]
    \centering
    \includegraphics[width=0.93\textwidth]{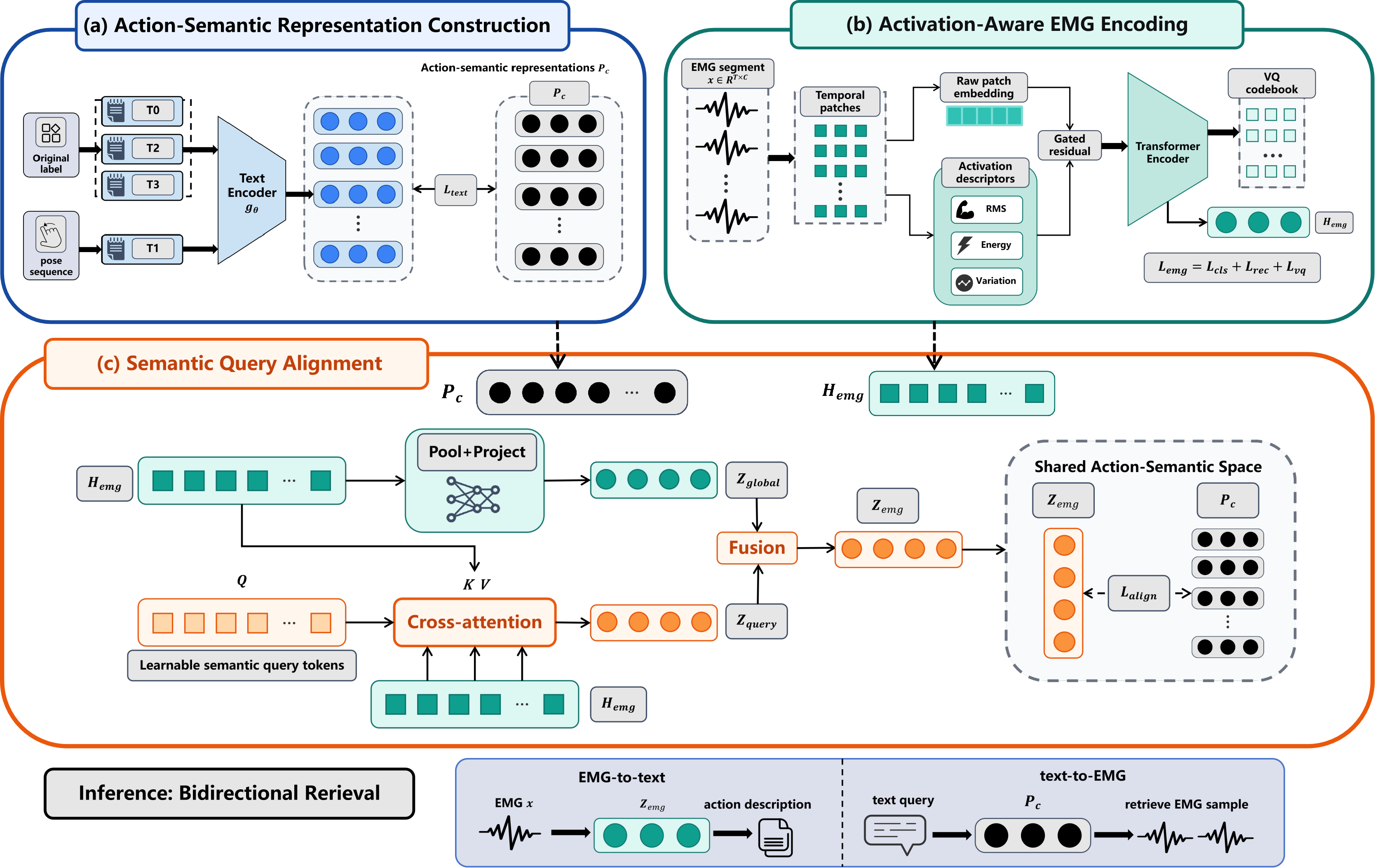}
    \caption{Overview of MyoSem. Multi-view descriptions organize the text space, an activation-aware encoder represents EMG segments, and semantic queries aggregate local EMG information for prototype alignment. The shared space supports mutual querying between EMG and action descriptions.}
    \label{fig:myosem_architecture}
\end{figure*}

\subsection{Task Formulation}

Following language-supervised retrieval and sensor--language grounding \cite{radford2021clip,moon2023imu2clip}, we evaluate two directions: EMG-to-Text (E2T) retrieval from an EMG segment to action prototypes and Text-to-EMG (T2E) retrieval from a text prototype to EMG samples. For $x\in\mathbb{R}^{T\times C}$, the normalized EMG embedding $z_{\mathrm{emg}}\in\mathbb{R}^{d}$ is scored against $P_c\in\mathbb{R}^{d}$ by cosine similarity:
\begin{equation}
    s(x,c)=z_{\mathrm{emg}}(x)^\top P_c.
\end{equation}
Cross-class training uses seen-class EMG and descriptions only; held-out class descriptions are encoded at inference to form target prototypes, giving inductive zero-shot retrieval.

\subsection{Multi-View Action-Semantic Text Space}

Short labels omit action process, participating fingers, and final configuration. We therefore represent each class with complementary views, following semantic label expansion in activity recognition \cite{yan2025lanhar}:
\begin{equation}
    \mathcal{T}_c=\{T_c^0,T_c^1,T_c^2,T_c^3\},
\end{equation}
Here $T_0$ is the class anchor, $T_1$ a structured process description, $T_2$ a canonical action and final-state description, and $T_3$ an alternative label-based phrasing when available.

$T_1$ combines dataset annotations, category labels, and Chinese templates \cite{atzori2014ninapro,salter2024emg2pose}. NinaPro uses class-and-stage templates, whereas EMG2Pose adds recording-level joint-angle summaries. Because NinaPro already provides class labels and canonical final-hand descriptions, we set $T_3=T_2$; EMG2Pose contains staged and compositional actions, so its $T_3$ preserves the action sequence in a compact paraphrase. The encoder is trained on Chinese descriptions; English text is translated for presentation. For $t_{c,n}^{(v)}$,
\begin{equation}
    e_{c,n}^{(v)}=\mathrm{norm}\bigl(g_\theta(t_{c,n}^{(v)})\bigr),
\end{equation}
we use a multi-positive contrastive objective \cite{oord2018cpc,khosla2020supcon}, with $\mathcal{P}(i)=\{j\mid y_j=y_i\}$ denoting batch positives and $\tau$ the text-space temperature. For views $(u,v)$,
\begin{equation}
\mathcal{L}_{u\rightarrow v}
=-\frac{1}{B}\sum_{i=1}^{B}
\log
\frac{\sum_{j\in\mathcal{P}(i)}\exp\!\left(\tau^{-1}\langle e_i^{(u)},e_j^{(v)}\rangle\right)}
{\sum_{j=1}^{B}\exp\!\left(\tau^{-1}\langle e_i^{(u)},e_j^{(v)}\rangle\right)},
\end{equation}
and $\mathcal{L}_{u\leftrightarrow v}=\tfrac{1}{2}(\mathcal{L}_{u\rightarrow v}+\mathcal{L}_{v\rightarrow u})$. The complete objective is
\vspace{-3pt}
\begin{equation}
\begin{aligned}
\mathcal{L}_{\mathrm{text}}
&=\mathcal{L}_{T_1\leftrightarrow T_2}
+\lambda_{02}\mathcal{L}_{T_0\leftrightarrow T_2}\\
&\quad+\lambda_{23}\mathcal{L}_{T_2\leftrightarrow T_3}
+\lambda_{13}\mathcal{L}_{T_1\leftrightarrow T_3}.
\end{aligned}
\end{equation}
The first two terms link process and canonical semantics while preserving the class anchor; the others maintain view consistency. $T_1$ supervises text-space training, and retrieval prototypes aggregate $T_0$, $T_2$, and $T_3$. With $\bar e_c^{(v)}$ the mean view embedding,
\begingroup
\setlength{\abovedisplayskip}{2pt}
\setlength{\belowdisplayskip}{1pt}
\begin{equation}
\begin{aligned}
\bar e_c^{(v)}
&=\mathrm{norm}\left(\frac{1}{N_c^{(v)}}\sum_{n=1}^{N_c^{(v)}}e_{c,n}^{(v)}\right),\\
P_c
&=\mathrm{norm}\left(\frac{1}{3}\sum_{v\in\{0,2,3\}}\bar e_c^{(v)}\right).
\end{aligned}
\end{equation}
\endgroup
$P_c$ provides the fixed semantic reference for EMG alignment and retrieval.

\subsection{Activation-Aware EMG Encoder}

\begin{figure*}[!t]
\centering
\begin{minipage}{.48\textwidth}
\centering
\includegraphics[width=\linewidth]{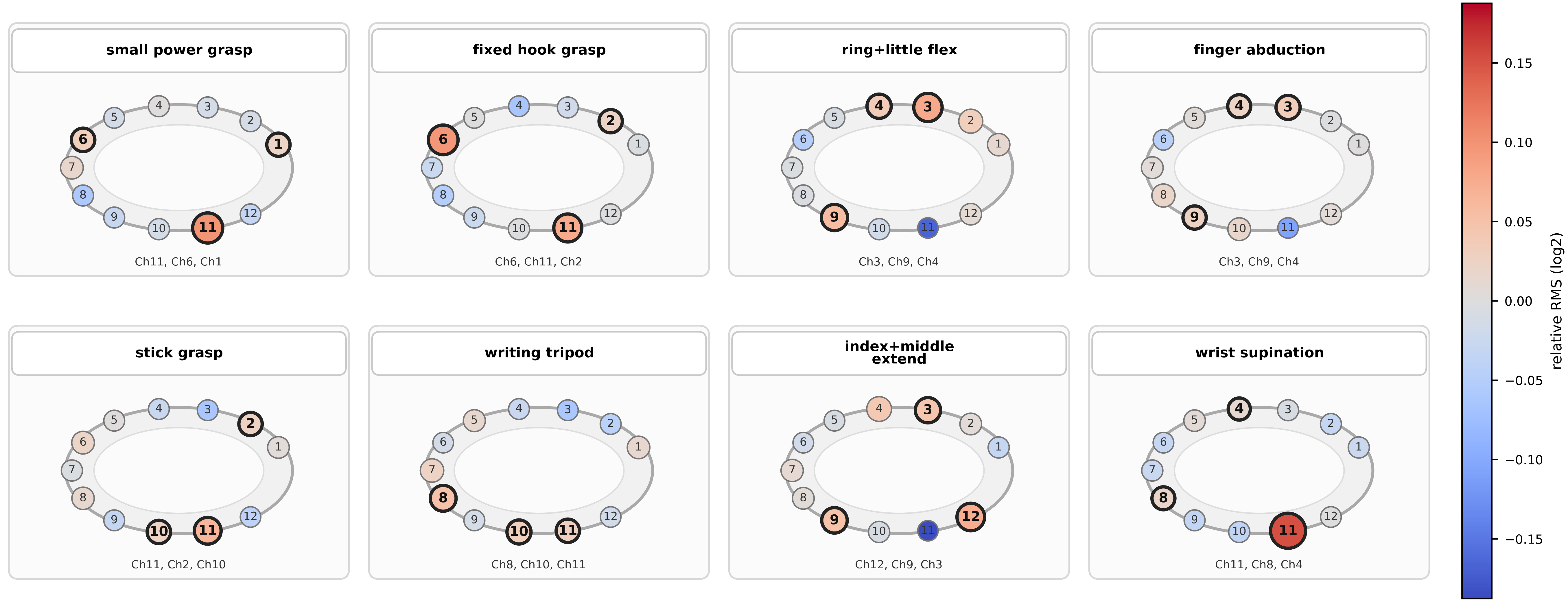}
\end{minipage}
\hfill
\begin{minipage}{.48\textwidth}
\centering
\includegraphics[width=\linewidth]{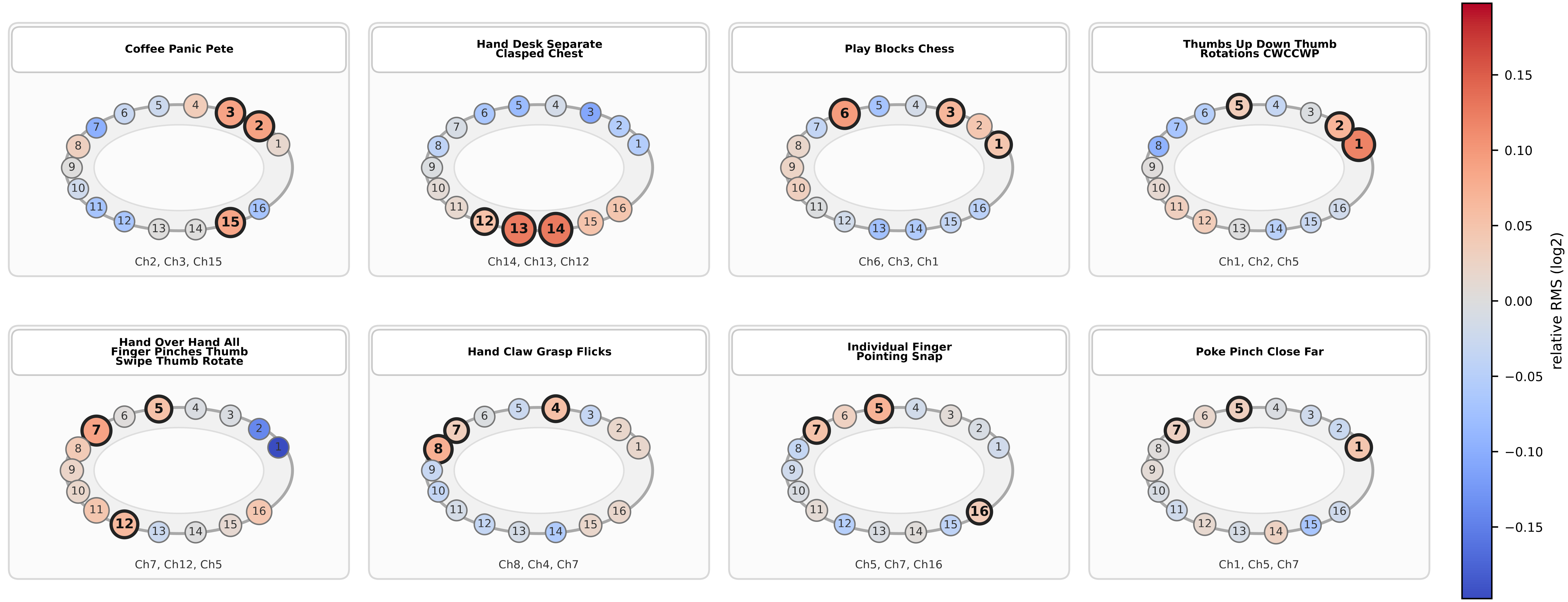}
\end{minipage}
\caption{Class-conditioned relative RMS activation on NinaPro (left) and EMG2Pose (right). Numbered circles are EMG channels; warmer colors indicate higher normalized RMS.}
\label{fig:channel_topology}
\end{figure*}

Patch embeddings may not retain amplitude and variation cues used in time-domain EMG analysis \cite{hudgins1993emgfeatures}. For a patch of length $p$ with raw patch embedding $r_t$, we inject local standard deviation (a centered-RMS measure), log mean-square energy, and mean absolute temporal variation:
\begin{equation}
\begin{aligned}
\mathrm{std}_t
&=\sqrt{\mathbb{E}_{p}[(x-\bar{x})^2]},\\
\mathrm{energy}_t
&=\log\bigl(1+\mathbb{E}_{p}[x^2]\bigr),\\
\mathrm{variation}_t
&=\mathbb{E}_{p}[|\Delta x|].
\end{aligned}
\end{equation}
For a $C$-channel patch, $\mathbf d_t^{e}=[\mathrm{std}_t;\mathrm{energy}_t]\in\mathbb{R}^{2C}$ and $\mathbf d_t^{v}=\mathrm{variation}_t\in\mathbb{R}^{C}$ enter gated residual branches:
\begin{equation}
h_t=r_t
+\sigma(g_e)\phi_e(\mathbf d_t^{e})
+\sigma(g_v)\phi_v(\mathbf d_t^{v}),
\end{equation}
Here $\phi_e,\phi_v$ are MLPs and $g_e,g_v$ are learnable scalar gates. Fig.~\ref{fig:channel_topology} visualizes action-dependent activation patterns; Table~\ref{tab:ablation} ablates the descriptor branches.

\begin{figure*}[t]
\centering
\includegraphics[width=\textwidth]{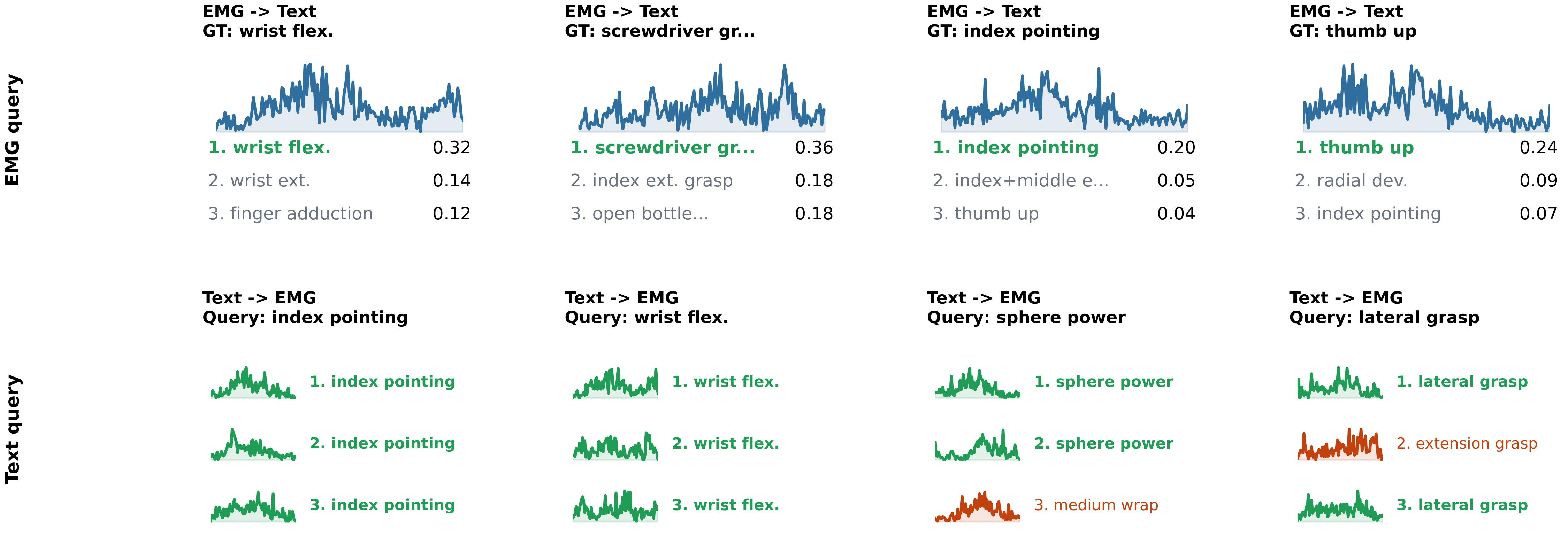}
\vspace{0.1em}

\fontsize{6.0}{6.4}\selectfont
\setlength{\tabcolsep}{1.8pt}
\begin{tabular}{@{}lp{0.43\textwidth}p{0.43\textwidth}@{}}
\toprule
View & NinaPro & EMG2Pose \\
\midrule
$T_0$ & Thumb up. & Coffee-cup grasp and toy squeeze. \\
$T_1$ & The thumb gradually extends while the other fingers move toward the palm, forming a stable thumb-up hand shape. & The action first resembles grasping a cup handle and then changes into a whole-hand squeezing motion. \\
$T_2$ & The thumb is extended upward while the other four fingers are flexed or closed. & The hand grasps a cup-like handle and then squeezes a soft object with the whole hand. \\
$T_3$ & The thumb is extended upward while the other four fingers are flexed or closed. & Hold a cup handle, then squeeze a soft toy. \\
\bottomrule
\end{tabular}
\caption{Bidirectional retrieval cases and representative multi-view descriptions. EMG-to-Text (E2T) ranks action prototypes, whereas Text-to-EMG (T2E) ranks gallery samples; green and orange denote matches and mismatches. English descriptions translate the Chinese training text; NinaPro $T_3$ reuses $T_2$.}
\label{fig:bidirectional_retrieval_cases}
\end{figure*}

We regularize the Transformer-encoded patch sequence $H=\{h_t\}$ \cite{vaswani2017attention} with a vector-quantized codebook \cite{oord2017vqvae}. For projected patch state $q_t$,
\vspace{-3pt}
\begin{equation}
    k^\ast=\arg\max_k
    \left\langle\mathrm{norm}(q_t),\mathrm{norm}(E_k)\right\rangle.
\end{equation}
Stage 2 combines action discrimination, reconstruction, and VQ regularization:
\vspace{-2pt}
\begin{equation}
    \mathcal{L}_{\mathrm{emg}}
    =\mathcal{L}_{\mathrm{cls}}
    +\lambda_{\mathrm{rec}}\mathcal{L}_{\mathrm{rec}}
    +\lambda_{\mathrm{vq}}\mathcal{L}_{\mathrm{vq}}.
\end{equation}
Here $\mathcal{L}_{\mathrm{cls}}$, $\mathcal{L}_{\mathrm{rec}}$, and $\mathcal{L}_{\mathrm{vq}}$ denote classification, reconstruction, and vector-quantization/codebook losses; $\lambda_{\mathrm{rec}}$ and $\lambda_{\mathrm{vq}}$ weight the latter two. Stage 3 uses the pre-VQ segment representation and patch sequence.

\subsection{Semantic Query Alignment}

Semantic query alignment uses learnable queries to aggregate local EMG evidence for prototype alignment \cite{li2023blip2}. Given $H_{\mathrm{emg}}\in\mathbb{R}^{L\times d}$, four learned queries first produce
\begin{equation}
    z_{\mathrm{global}}
    =\mathrm{norm}\bigl(f_{\mathrm{proj}}(\mathrm{Pool}(H_{\mathrm{emg}}))\bigr).
\end{equation}
With learned global queries $Q^g=\{q_j^g\}_{j=1}^{4}$, the pooling operator is
\begin{equation}
    \mathrm{Pool}(H_{\mathrm{emg}})
    =\frac{1}{4}\sum_{j=1}^{4}
    \mathrm{Attn}(q_j^g,H_{\mathrm{emg}},H_{\mathrm{emg}}).
\end{equation}
Then $M$ semantic queries $Q\in\mathbb{R}^{M\times d}$ attend to EMG patch states:
\begin{equation}
\begin{aligned}
\widetilde{Q}
&=\mathrm{Attn}(Q,H_{\mathrm{emg}},H_{\mathrm{emg}}),\\
z_{\mathrm{query}}
&=\mathrm{norm}\left(\frac{1}{M}\sum_{m=1}^{M}f_q(\widetilde{Q}_m)\right).
\end{aligned}
\end{equation}
The final representation uses fixed $\alpha=0.5$:
\begin{equation}
    z_{\mathrm{emg}}
    =\mathrm{norm}\bigl((1-\alpha)z_{\mathrm{global}}+\alpha z_{\mathrm{query}}\bigr).
\end{equation}
For $(x_i,y_i)$, frozen prototypes provide alignment targets:
\enlargethispage{-4pt}
\begingroup
\setlength{\abovedisplayskip}{3pt}
\setlength{\belowdisplayskip}{2pt}
\begin{equation}
\mathcal{L}_{\mathrm{align}}
=-\frac{1}{B}\sum_{i=1}^{B}
\log
\frac{\exp\!\left(\tau_a^{-1}z_i^\top P_{y_i}\right)}
{\sum_{c\in\mathcal{C}}\exp\!\left(\tau_a^{-1}z_i^\top P_c\right)}.
\end{equation}
\endgroup
Here $\tau_a$ is the alignment temperature controlling prototype-logit sharpness.
An auxiliary classification loss is used during training:
\begingroup
\setlength{\abovedisplayskip}{2pt}
\setlength{\belowdisplayskip}{2pt}
\begin{equation}
    \mathcal{L}_{\mathrm{stage3}}
    =\mathcal{L}_{\mathrm{align}}
    +\lambda_{\mathrm{aux}}\mathcal{L}_{\mathrm{aux}}.
\end{equation}
\endgroup
Stage 1 learns $P_c$; Stage 2 trains the EMG encoder. Stage 3 freezes text representations, trains the projection and queries, then fine-tunes six EMG Transformer blocks and pooling. At inference, EMG-to-text ranks prototypes by $z_{\mathrm{emg}}^\top P_c$; text-to-EMG ranks EMG embeddings by $P_c$.

\section{Experiments}
\label{sec:experiments}

\newcommand{\rrcell}[2]{\shortstack{\textit{R1}\; $#1$\\[-1pt]\textit{R3}\; $#2$}}

\begin{figure*}[t]
\centering
\begin{minipage}{.495\textwidth}
\centering
\includegraphics[width=\linewidth]{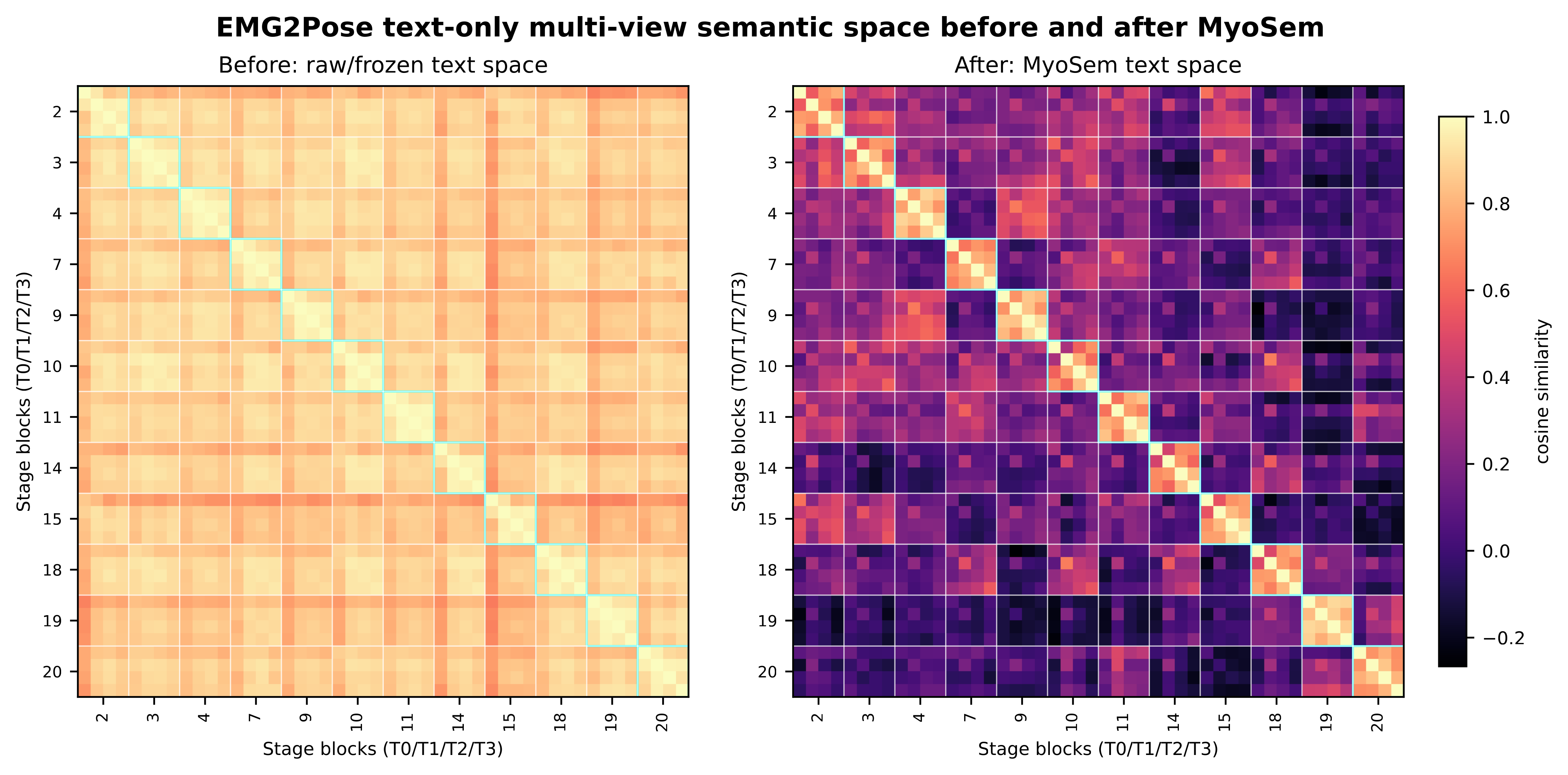}
\end{minipage}
\hfill
\begin{minipage}{.495\textwidth}
\centering
\includegraphics[width=\linewidth]{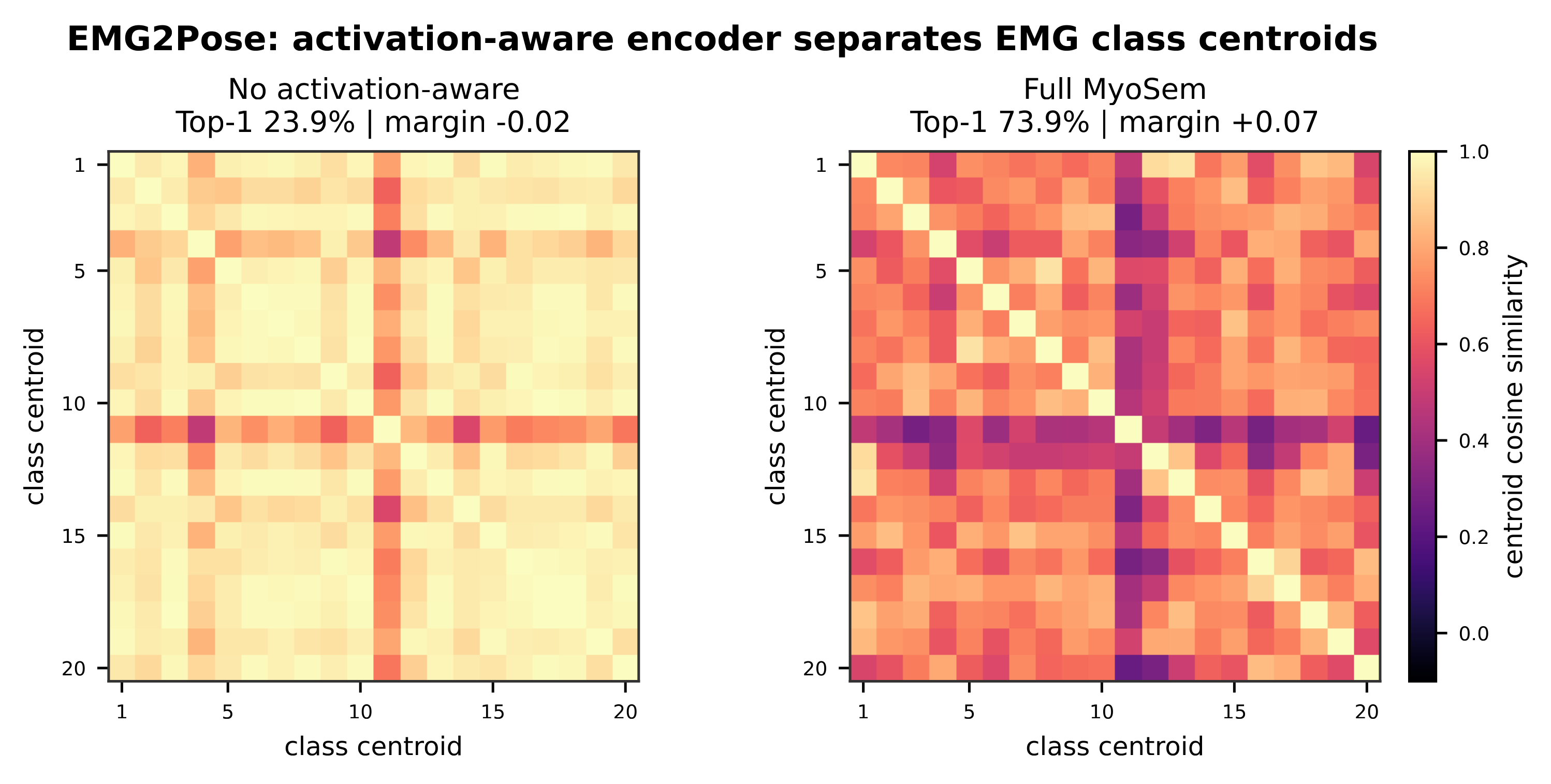}
\end{minipage}
\vspace{0.6em}

\begin{minipage}{.495\textwidth}
\centering
\includegraphics[width=\linewidth]{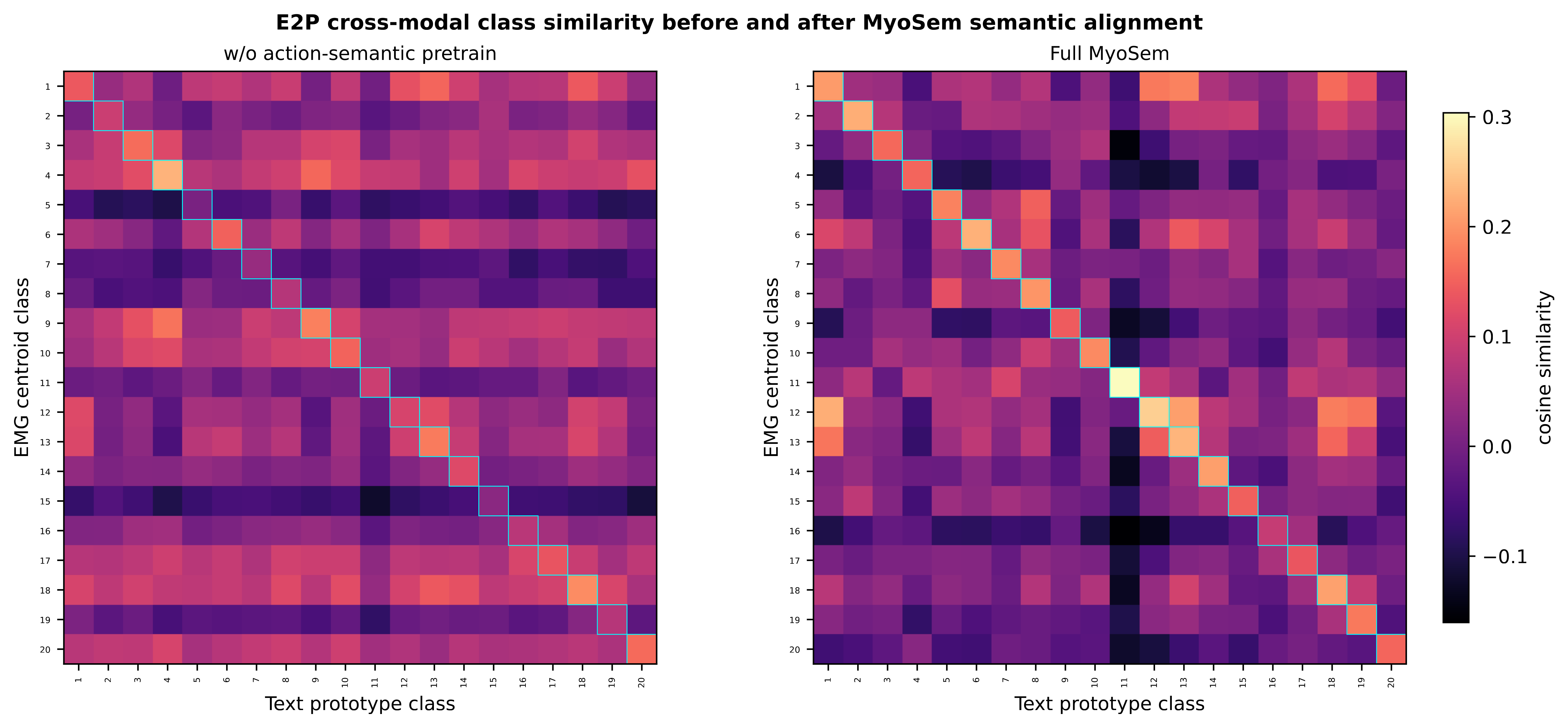}
\end{minipage}
\hfill
\begin{minipage}{.495\textwidth}
\centering
\includegraphics[width=\linewidth]{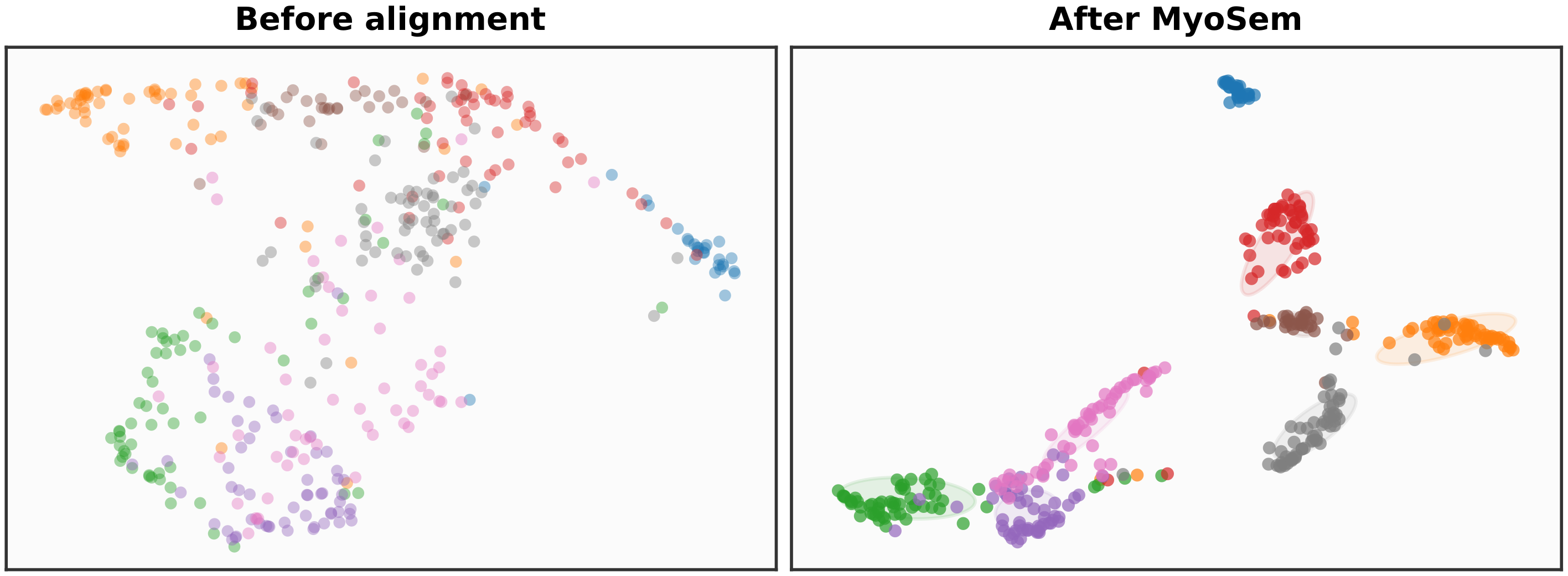}
\end{minipage}
\caption{Learned action-semantic alignment on EMG2Pose. Panels (a)--(c) use all 20 action classes; centroids use all available samples. Panel (d) projects samples from eight representative classes.}
\label{fig:semantic_alignment_analysis}
\end{figure*}

\makeatletter
\newcommand{\renderninaprofigure}{%
\begin{minipage}{\textwidth}
\centering
\begin{minipage}{.495\textwidth}
\centering
\includegraphics[width=\linewidth]{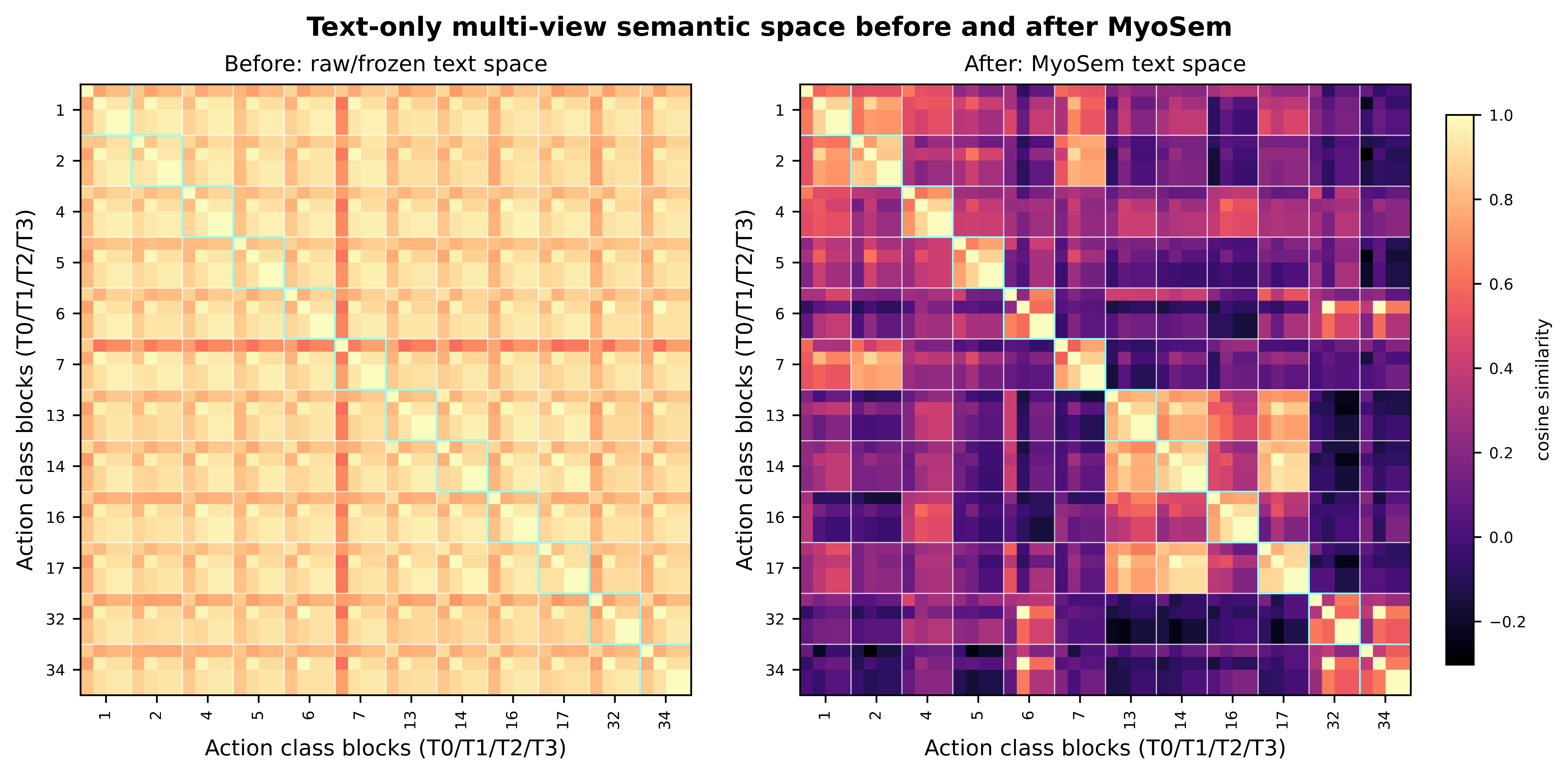}
\end{minipage}
\hfill
\begin{minipage}{.495\textwidth}
\centering
\includegraphics[width=\linewidth]{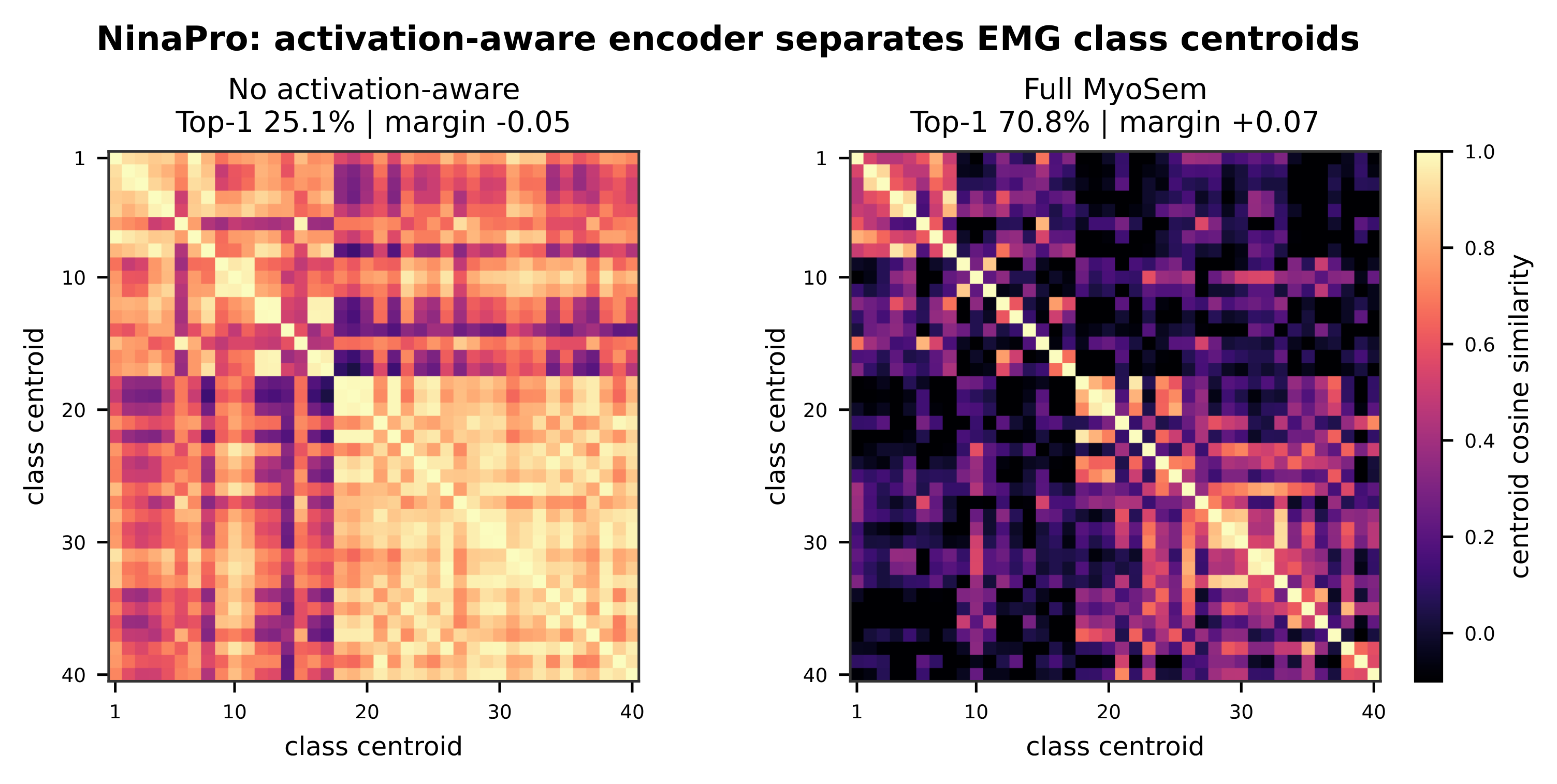}
\end{minipage}
\vspace{0.6em}

\begin{minipage}{.495\textwidth}
\centering
\includegraphics[width=\linewidth]{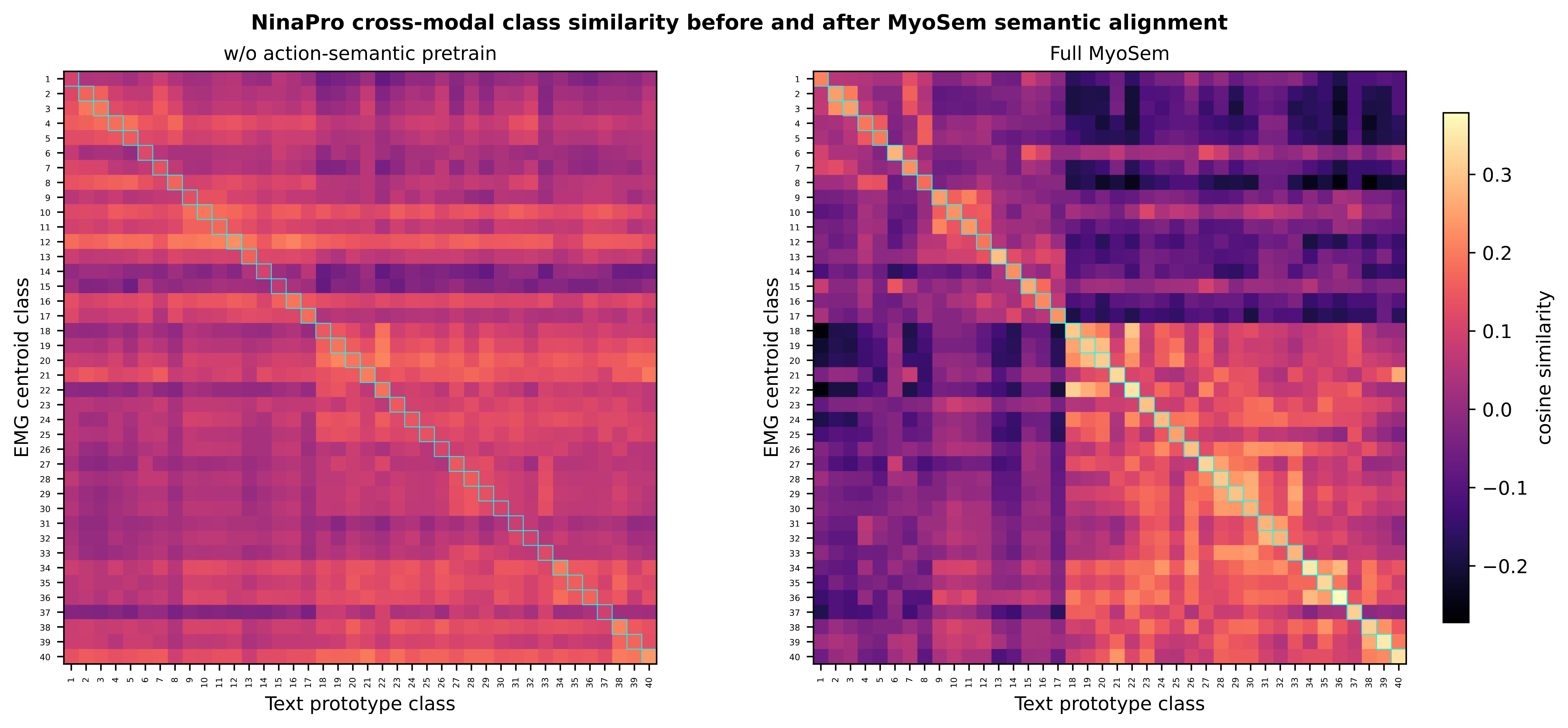}
\end{minipage}
\hfill
\begin{minipage}{.495\textwidth}
\centering
\includegraphics[width=\linewidth]{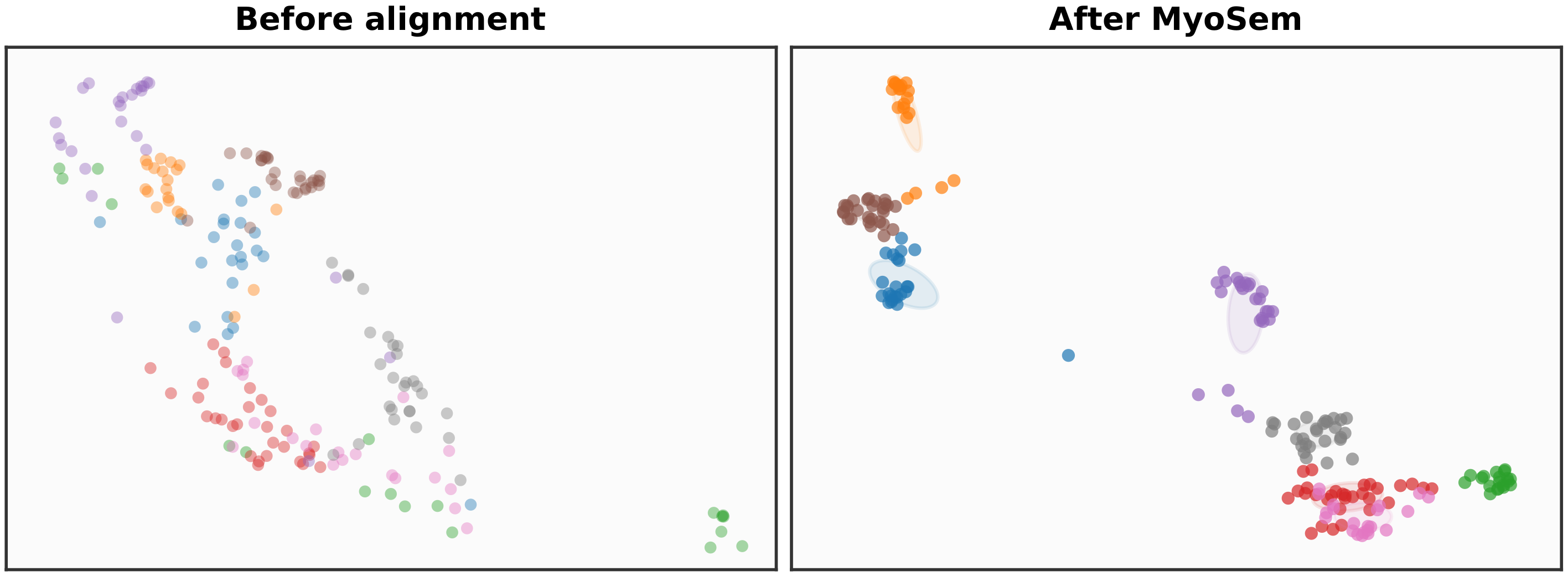}
\end{minipage}
\def\@captype{figure}%
\caption{Learned action-semantic alignment on NinaPro. Panels (a)--(c) use all 40 action classes; centroids use all available samples. Panel (d) projects samples from eight representative classes.}
\label{fig:ninapro_alignment_analysis}
\end{minipage}%
}
\makeatother

\begin{table}[!t]
\caption{Protocol composition. Entries give classes/users (C/U); Cand. is the final-test candidate count.}
\label{tab:protocol_composition}
\centering
\scriptsize
\setlength{\tabcolsep}{2.8pt}
\renewcommand{\arraystretch}{0.88}
\begin{tabular}{@{}lcccc@{}}
\toprule
Protocol & Train & Val. view & Test & Cand. \\
\midrule
DB2 Rand. & 40/40 & 40/40 & 40/40 & 40 \\
DB2 User  & 40/30 & 40/30 & 40/10 & 40 \\
DB2 Class & 30/40 & 30/40 & 10/40 & 10 \\
E2P Rand. & 20/158 & 20/157 & 20/157 & 20 \\
E2P User  & 20/158 & 20/15 & 20/20 & 20 \\
E2P Class & 20/158 & 20/143 & 5/158 & 5 \\
DB2$\rightarrow$DB3 & 40/40 & 40/40 & 8/11 & 8 \\
\bottomrule
\end{tabular}
\end{table}

\begin{table*}[!t]
\centering
\caption{EMG-to-text retrieval (\%). Cells report R@1/R@3; $\pm$ is bootstrap standard deviation over test samples. Best values are bolded.}
\label{tab:e2t_main}
\fontsize{6.7}{7.6}\selectfont
\setlength{\tabcolsep}{2.2pt}
\begin{tabular*}{\textwidth}{@{\extracolsep{\fill}}lccccccc@{}}
\toprule
& \multicolumn{3}{c}{NinaPro DB2} & \multicolumn{3}{c}{EMG2Pose} & DB2$\rightarrow$DB3 \\
\cmidrule(lr){2-4}\cmidrule(lr){5-7}\cmidrule(l){8-8}
Model & Rand. & User & Class & Rand. & User & Class & Transfer \\
\midrule
TCN~\cite{bai2018tcn}
& \rrcell{51.15{\pm}1.57}{80.38{\pm}1.24}
& \rrcell{32.71{\pm}0.95}{58.13{\pm}1.02}
& \rrcell{15.13{\pm}0.72}{44.87{\pm}0.98}
& \rrcell{46.59{\pm}1.61}{71.34{\pm}1.42}
& \rrcell{52.51{\pm}1.33}{78.65{\pm}1.11}
& \rrcell{30.72{\pm}1.05}{63.68{\pm}1.11}
& \rrcell{19.77{\pm}1.79}{45.54{\pm}2.17} \\
Attn-CNN-RNN~\cite{hu2018attncnnrnn}
& \rrcell{47.55{\pm}1.60}{75.98{\pm}1.35}
& \rrcell{32.00{\pm}0.95}{58.96{\pm}1.04}
& \rrcell{14.88{\pm}0.73}{46.96{\pm}1.04}
& \rrcell{34.07{\pm}1.49}{60.22{\pm}1.56}
& \rrcell{30.53{\pm}1.36}{56.28{\pm}1.41}
& \rrcell{33.46{\pm}1.13}{58.42{\pm}1.15}
& \rrcell{20.54{\pm}1.82}{45.35{\pm}2.26} \\
CNN-BiLSTM~\cite{lecun1998lenet,hochreiter1997lstm,schuster1997birnn}
& \rrcell{41.44{\pm}1.62}{72.97{\pm}1.43}
& \rrcell{34.29{\pm}0.97}{56.92{\pm}0.98}
& \rrcell{10.00{\pm}0.60}{30.63{\pm}0.91}
& \rrcell{36.37{\pm}1.49}{61.82{\pm}1.52}
& \rrcell{37.68{\pm}1.40}{65.15{\pm}1.38}
& \rrcell{34.30{\pm}1.13}{63.35{\pm}1.11}
& \rrcell{16.86{\pm}1.65}{47.87{\pm}2.17} \\
SE-CNN~\cite{xu2023secnn}
& \rrcell{34.83{\pm}1.43}{61.06{\pm}1.53}
& \rrcell{22.62{\pm}0.82}{42.58{\pm}0.98}
& \rrcell{10.00{\pm}0.61}{30.00{\pm}0.92}
& \rrcell{36.57{\pm}1.48}{60.72{\pm}1.53}
& \rrcell{41.92{\pm}1.43}{67.82{\pm}1.31}
& \rrcell{32.85{\pm}1.13}{63.57{\pm}1.12}
& \rrcell{18.99{\pm}1.73}{46.12{\pm}2.14} \\
Atzori CNN~\cite{atzori2016cnn}
& \rrcell{32.23{\pm}1.38}{58.36{\pm}1.55}
& \rrcell{23.42{\pm}0.84}{46.04{\pm}0.97}
& \rrcell{16.50{\pm}0.78}{49.63{\pm}1.02}
& \rrcell{23.45{\pm}1.36}{43.89{\pm}1.62}
& \rrcell{28.96{\pm}1.25}{53.22{\pm}1.38}
& \rrcell{24.79{\pm}0.99}{71.12{\pm}1.05}
& \rrcell{15.31{\pm}1.57}{46.12{\pm}2.21} \\
\midrule
CPEP~\cite{cui2025cpep}
& \rrcell{57.56{\pm}1.51}{86.19{\pm}1.10}
& \rrcell{37.25{\pm}0.97}{63.00{\pm}1.00}
& \rrcell{\mathbf{33.04{\pm}0.96}}{67.92{\pm}0.97}
& \rrcell{31.66{\pm}1.46}{58.62{\pm}1.53}
& \rrcell{36.26{\pm}1.41}{61.70{\pm}1.42}
& \rrcell{26.97{\pm}1.08}{62.95{\pm}1.11}
& \rrcell{18.22{\pm}1.71}{45.74{\pm}2.19} \\
EMBridge~\cite{cui2026embridge}
& \rrcell{58.76{\pm}1.55}{86.69{\pm}1.06}
& \rrcell{37.75{\pm}1.01}{64.13{\pm}1.01}
& \rrcell{32.62{\pm}0.97}{66.25{\pm}0.95}
& \rrcell{33.77{\pm}1.51}{58.02{\pm}1.60}
& \rrcell{33.67{\pm}1.34}{59.73{\pm}1.39}
& \rrcell{32.40{\pm}1.15}{65.36{\pm}1.07}
& \rrcell{\mathbf{21.12{\pm}1.82}}{46.51{\pm}2.22} \\
AEMG~\cite{huang2026aemg}
& \rrcell{24.62{\pm}1.38}{50.85{\pm}1.58}
& \rrcell{16.75{\pm}0.74}{36.33{\pm}0.96}
& \rrcell{18.92{\pm}0.80}{49.88{\pm}1.03}
& \rrcell{10.62{\pm}1.00}{27.96{\pm}1.40}
& \rrcell{14.21{\pm}1.02}{36.42{\pm}1.35}
& \rrcell{26.92{\pm}1.02}{64.69{\pm}1.15}
& \rrcell{17.05{\pm}1.72}{44.38{\pm}2.18} \\
\midrule
MyoSem
& \rrcell{\mathbf{69.77{\pm}1.41}}{\mathbf{92.99{\pm}0.79}}
& \rrcell{\mathbf{41.96{\pm}1.01}}{\mathbf{67.08{\pm}0.97}}
& \rrcell{29.37{\pm}0.93}{\mathbf{68.04{\pm}0.96}}
& \rrcell{\mathbf{76.05{\pm}1.34}}{\mathbf{90.88{\pm}0.91}}
& \rrcell{\mathbf{74.88{\pm}1.21}}{\mathbf{91.76{\pm}0.77}}
& \rrcell{\mathbf{37.83{\pm}1.16}}{\mathbf{77.39{\pm}0.99}}
& \rrcell{\mathbf{21.12{\pm}1.83}}{\mathbf{48.26{\pm}2.25}} \\
\bottomrule
\end{tabular*}
\end{table*}

\begin{figure*}[p]
\renderninaprofigure
\end{figure*}

\begin{table*}[p]
\centering
\caption{Text-to-EMG retrieval over 100 balanced-gallery episodes (\%). Cells report R@1/R@3; $\pm$ is standard error across episodes. Best values are bolded.}
\label{tab:t2e_main}
\fontsize{6.7}{7.6}\selectfont
\setlength{\tabcolsep}{2.2pt}
\begin{tabular*}{\textwidth}{@{\extracolsep{\fill}}lccccccc@{}}
\toprule
& \multicolumn{3}{c}{NinaPro DB2} & \multicolumn{3}{c}{EMG2Pose} & DB2$\rightarrow$DB3 \\
\cmidrule(lr){2-4}\cmidrule(lr){5-7}\cmidrule(l){8-8}
Model & Rand. & User & Class & Rand. & User & Class & Transfer \\
\midrule
BERT~\cite{devlin2019bert}
& \rrcell{5.93{\pm}0.28}{14.55{\pm}0.36}
& \rrcell{6.50{\pm}0.26}{16.18{\pm}0.37}
& \rrcell{12.10{\pm}0.82}{38.60{\pm}0.91}
& \rrcell{18.05{\pm}0.48}{40.85{\pm}0.69}
& \rrcell{16.90{\pm}0.54}{40.15{\pm}0.77}
& \rrcell{33.00{\pm}1.24}{70.60{\pm}1.43}
& \rrcell{14.37{\pm}0.85}{40.00{\pm}1.08} \\
MacBERT~\cite{cui2020macbert}
& \rrcell{3.50{\pm}0.15}{10.30{\pm}0.22}
& \rrcell{4.83{\pm}0.21}{13.55{\pm}0.29}
& \rrcell{10.30{\pm}0.33}{32.10{\pm}0.45}
& \rrcell{6.75{\pm}0.29}{19.10{\pm}0.45}
& \rrcell{8.55{\pm}0.36}{19.30{\pm}0.41}
& \rrcell{22.20{\pm}0.69}{61.60{\pm}0.67}
& \rrcell{12.62{\pm}0.25}{37.88{\pm}0.21} \\
CLIP~\cite{radford2021clip}
& \rrcell{10.35{\pm}0.37}{23.13{\pm}0.42}
& \rrcell{6.58{\pm}0.25}{16.45{\pm}0.32}
& \rrcell{9.40{\pm}0.66}{28.60{\pm}0.89}
& \rrcell{26.60{\pm}0.73}{48.30{\pm}0.78}
& \rrcell{26.60{\pm}0.81}{48.45{\pm}0.75}
& \rrcell{24.00{\pm}1.62}{71.20{\pm}1.48}
& \rrcell{17.75{\pm}1.45}{43.62{\pm}1.77} \\
Chinese-CLIP~\cite{yang2022chineseclip}
& \rrcell{12.15{\pm}0.49}{32.03{\pm}0.57}
& \rrcell{9.90{\pm}0.38}{25.62{\pm}0.55}
& \rrcell{17.00{\pm}0.83}{46.30{\pm}1.20}
& \rrcell{63.05{\pm}0.91}{84.30{\pm}0.72}
& \rrcell{62.80{\pm}0.97}{84.50{\pm}0.84}
& \rrcell{30.80{\pm}1.82}{73.60{\pm}1.52}
& \rrcell{19.12{\pm}1.34}{40.62{\pm}1.78} \\
\midrule
MyoSem
& \rrcell{\mathbf{57.65{\pm}0.67}}{\mathbf{85.75{\pm}0.52}}
& \rrcell{\mathbf{40.73{\pm}0.83}}{\mathbf{64.80{\pm}0.71}}
& \rrcell{\mathbf{35.20{\pm}1.09}}{\mathbf{71.60{\pm}1.14}}
& \rrcell{\mathbf{67.65{\pm}0.99}}{\mathbf{86.20{\pm}0.64}}
& \rrcell{\mathbf{66.35{\pm}1.01}}{\mathbf{86.20{\pm}0.68}}
& \rrcell{\mathbf{46.20{\pm}1.97}}{\mathbf{85.20{\pm}1.49}}
& \rrcell{\mathbf{23.88{\pm}1.54}}{\mathbf{47.75{\pm}1.64}} \\
\bottomrule
\end{tabular*}
\end{table*}

\begin{table*}[p]
\centering
\caption{Ablation for both retrieval directions (\%). Cells report R@1/R@3; uncertainty follows Tables~\ref{tab:e2t_main} and~\ref{tab:t2e_main}. Best values are bolded.}
\label{tab:ablation}
\fontsize{6.8}{7.7}\selectfont
\setlength{\tabcolsep}{2.2pt}
\begin{tabular*}{\textwidth}{@{\extracolsep{\fill}}lccccccc@{}}
\toprule
& \multicolumn{3}{c}{NinaPro DB2} & \multicolumn{3}{c}{EMG2Pose} & DB2$\rightarrow$DB3 \\
\cmidrule(lr){2-4}\cmidrule(lr){5-7}\cmidrule(l){8-8}
Variant & Rand. & User & Class & Rand. & User & Class & Transfer \\
\midrule
\multicolumn{8}{l}{\textit{EMG-to-text}} \\
w/o activation-aware
& \rrcell{23.92{\pm}1.33}{48.15{\pm}1.58} & \rrcell{23.46{\pm}0.84}{47.92{\pm}1.00} & \rrcell{20.83{\pm}0.82}{54.50{\pm}1.03}
& \rrcell{24.25{\pm}1.32}{49.20{\pm}1.55} & \rrcell{24.96{\pm}1.23}{47.41{\pm}1.40} & \rrcell{29.83{\pm}1.11}{55.01{\pm}1.16} & \rrcell{20.93{\pm}1.82}{43.80{\pm}2.20} \\
w/o semantic queries
& \rrcell{52.65{\pm}1.57}{82.58{\pm}1.17} & \rrcell{36.92{\pm}0.97}{63.29{\pm}0.98} & \rrcell{30.67{\pm}0.93}{65.71{\pm}0.98}
& \rrcell{69.94{\pm}1.42}{87.47{\pm}1.04} & \rrcell{70.41{\pm}1.25}{88.93{\pm}0.85} & \rrcell{33.97{\pm}1.14}{62.79{\pm}1.13} & \rrcell{19.57{\pm}1.80}{42.64{\pm}2.19} \\
w/o action-sem. pretrain
& \rrcell{46.35{\pm}1.56}{77.18{\pm}1.28} & \rrcell{30.04{\pm}0.94}{57.00{\pm}1.01} & \rrcell{\mathbf{32.08{\pm}0.93}}{65.96{\pm}0.95}
& \rrcell{65.33{\pm}1.45}{87.27{\pm}1.05} & \rrcell{66.17{\pm}1.33}{84.54{\pm}1.00} & \rrcell{22.61{\pm}0.95}{63.91{\pm}1.17} & \rrcell{18.60{\pm}1.70}{42.83{\pm}2.07} \\
Full MyoSem
& \rrcell{\mathbf{69.77{\pm}1.41}}{\mathbf{92.99{\pm}0.79}} & \rrcell{\mathbf{41.96{\pm}1.01}}{\mathbf{67.08{\pm}0.97}} & \rrcell{29.37{\pm}0.93}{\mathbf{68.04{\pm}0.96}}
& \rrcell{\mathbf{76.05{\pm}1.34}}{\mathbf{90.88{\pm}0.91}} & \rrcell{\mathbf{74.88{\pm}1.21}}{\mathbf{91.76{\pm}0.77}} & \rrcell{\mathbf{37.83{\pm}1.16}}{\mathbf{77.39{\pm}0.99}} & \rrcell{\mathbf{21.12{\pm}1.83}}{\mathbf{48.26{\pm}2.25}} \\
\midrule
\multicolumn{8}{l}{\textit{Text-to-EMG}} \\
w/o activation-aware
& \rrcell{19.53{\pm}0.49}{44.00{\pm}0.58} & \rrcell{19.53{\pm}0.53}{42.32{\pm}0.68} & \rrcell{21.20{\pm}1.00}{54.50{\pm}1.35}
& \rrcell{13.75{\pm}0.56}{32.50{\pm}0.71} & \rrcell{14.20{\pm}0.60}{34.40{\pm}0.79} & \rrcell{25.40{\pm}1.49}{66.40{\pm}1.85} & \rrcell{17.62{\pm}1.21}{42.62{\pm}1.63} \\
w/o semantic queries
& \rrcell{41.73{\pm}0.54}{74.38{\pm}0.61} & \rrcell{32.93{\pm}0.59}{57.42{\pm}0.62} & \rrcell{30.40{\pm}1.18}{62.10{\pm}1.19}
& \rrcell{59.60{\pm}1.03}{82.00{\pm}0.75} & \rrcell{59.00{\pm}0.97}{81.90{\pm}0.70} & \rrcell{43.80{\pm}1.91}{82.20{\pm}1.44} & \rrcell{15.00{\pm}1.17}{44.00{\pm}1.42} \\
w/o action-sem. pretrain
& \rrcell{16.63{\pm}0.51}{36.30{\pm}0.64} & \rrcell{13.43{\pm}0.52}{29.07{\pm}0.61} & \rrcell{20.70{\pm}0.93}{50.70{\pm}1.22}
& \rrcell{25.75{\pm}0.91}{50.10{\pm}0.89} & \rrcell{25.45{\pm}0.92}{49.70{\pm}0.93} & \rrcell{23.00{\pm}0.77}{68.60{\pm}1.39} & \rrcell{13.63{\pm}0.87}{39.62{\pm}1.13} \\
Full MyoSem
& \rrcell{\mathbf{57.65{\pm}0.67}}{\mathbf{85.75{\pm}0.52}} & \rrcell{\mathbf{40.73{\pm}0.83}}{\mathbf{64.80{\pm}0.71}} & \rrcell{\mathbf{35.20{\pm}1.09}}{\mathbf{71.60{\pm}1.14}}
& \rrcell{\mathbf{67.65{\pm}0.99}}{\mathbf{86.20{\pm}0.64}} & \rrcell{\mathbf{66.35{\pm}1.01}}{\mathbf{86.20{\pm}0.68}} & \rrcell{\mathbf{46.20{\pm}1.97}}{\mathbf{85.20{\pm}1.49}} & \rrcell{\mathbf{23.88{\pm}1.54}}{\mathbf{47.75{\pm}1.64}} \\
\bottomrule
\end{tabular*}
\end{table*}

\subsection{Experimental Protocol}

Seven protocols cover random, cross-user, and cross-class NinaPro DB2/EMG2Pose (E2P) splits and eight-class DB2$\rightarrow$DB3 transfer \cite{atzori2014ninapro,salter2024emg2pose}. They examine EMG--description correspondence in-distribution, across user shifts, for held-out action classes, and across able-bodied-to-amputee populations. User and class protocols enforce $\mathcal U_{\mathrm{tr}}\cap\mathcal U_{\mathrm{te}}=\varnothing$ or $\mathcal C_{\mathrm{tr}}\cap\mathcal C_{\mathrm{te}}=\varnothing$; held-out descriptions are encoded only at inference.

Table~\ref{tab:protocol_composition} summarizes class/user composition. NinaPro cross-class and E2P cross-class select checkpoints on their seen-class views; E2P cross-user uses the E2P random-split checkpoint. For DB2$\rightarrow$DB3, pooled DB2 partitions provide source training, DB2 validation selects checkpoints, and the fixed eight-grasp DB3 set is reserved for final evaluation with $\alpha_{\mathrm{text}}=0.94$.

Using $s(x,c)$ from Section~\ref{sec:method}, E2T ranks candidate prototypes for an EMG segment $x$:
\begingroup
\setlength{\abovedisplayskip}{3pt}
\setlength{\belowdisplayskip}{3pt}
\begin{equation}
\mathrm{E2T}(x)=\operatorname{rank}_{c\in\mathcal C_{\mathrm{te}}}s(x,c).
\end{equation}
\endgroup
T2E ranks a balanced gallery $\mathcal G$ with one test EMG segment per class:
\begingroup
\setlength{\abovedisplayskip}{3pt}
\setlength{\belowdisplayskip}{3pt}
\begin{equation}
\mathrm{T2E}(c)=\operatorname{rank}_{x\in\mathcal G}s(x,c).
\end{equation}
\endgroup
We repeat T2E over 100 independent galleries and report the mean and standard error. Recall is
\begin{equation}
\mathrm{R@K}=\frac{1}{N}\sum_i\mathbb I[r_i\le K],\qquad K\in\{1,3\}.
\end{equation}
For E2T, $\pm$ denotes bootstrap variability; for T2E, it denotes the standard error across 100 galleries. Conventional EMG networks, CPEP, EMBridge, and AEMG \cite{bai2018tcn,hu2018attncnnrnn,lecun1998lenet,hochreiter1997lstm,schuster1997birnn,xu2023secnn,atzori2016cnn,cui2025cpep,cui2026embridge,huang2026aemg} map segments to the same frozen prototypes. T2E text baselines use adapters to map descriptions into the fixed EMG space and balanced galleries \cite{devlin2019bert,cui2020macbert,radford2021clip,yang2022chineseclip}; all baselines use matched splits and candidates. We use eight semantic queries and initialize $g_e=g_v=-4.0$ (sigmoid $\approx0.018$).

\subsection{Distribution-Shift Retrieval}

E2T and T2E evaluate prototype identification and EMG evidence retrieval, respectively. MyoSem attains the best E2T R@3 in all seven protocols and the best R@1 in five, tying EMBridge on DB2$\rightarrow$DB3 R@1; on NinaPro cross-class E2T, CPEP leads R@1 while MyoSem leads R@3 (Table~\ref{tab:e2t_main}). MyoSem leads all T2E metrics in Table~\ref{tab:t2e_main}.

\subsection{Component Contributions}

Ablations remove activation-descriptor residuals, replace semantic queries with global pooling, and construct prototypes with frozen MacBERT, isolating local physiological evidence, local-evidence aggregation, and the learned action-semantic reference. The full model leads in every T2E and most E2T metrics (Table~\ref{tab:ablation}); activation descriptors have the largest overall effect.

\subsection{Learned Alignment Structure}

Figs.~\ref{fig:semantic_alignment_analysis} and~\ref{fig:ninapro_alignment_analysis} inspect the learned correspondence. Panel (a) compares text views with $T_{(c,u),(c',v)}=(e_c^{(u)})^\top e_{c'}^{(v)}$. Panel (b) uses the normalized EMG class centroid
\begin{equation}
\mu_c=\operatorname{norm}\!\left(\frac{1}{|\mathcal D_c|}\sum_{x\in\mathcal D_c}z_x\right),
\end{equation}
to form $A_{cc'}=\mu_c^\top\mu_{c'}$ for EMG-class separation. Panel (c) uses $S_{cc'}=\mu_c^\top P_{c'}$ for EMG--text correspondence; panel (d) projects $q_x=[s(x,1),\ldots,s(x,|\mathcal C|)]$ with UMAP \cite{mcinnes2018umap}. Table~\ref{tab:ablation} quantifies component contributions, and Fig.~\ref{fig:bidirectional_retrieval_cases} shows representative retrievals.

\section{Conclusion}
\label{sec:conclusion}

MyoSem aligns EMG and multi-view action descriptions through semantic organization, activation-aware encoding, and semantic queries. Across seven NinaPro DB2, EMG2Pose, and DB2$\rightarrow$DB3 protocols, it achieves the highest E2T R@3 in all seven protocols and R@1 in five, ties DB2$\rightarrow$DB3 R@1, and leads all evaluated T2E text encoders. The results support bidirectional retrieval across in-distribution, user, class, and population settings.

\bibliographystyle{IEEEtran}
\bibliography{bib/references}

\end{document}